\documentclass[10pt,letterpaper,twocolumn]{article}

\usepackage[margin=1.9cm]{geometry}

\usepackage{hyperref}
\usepackage{graphicx}
\usepackage{caption}
\usepackage{subcaption}

\usepackage[none]{hyphenat}

\usepackage{xcolor}

\usepackage{listings}
\usepackage{enumitem}

\usepackage{lipsum}
\usepackage{cuted}

\begin{document}

\twocolumn[

\title{Collective Knowledge: organizing research projects 
as a database of reusable components and portable workflows with common APIs}

\author{Grigori Fursin \\
\\
cTuning foundation and cKnowledge SAS
\\
\\
\href{https://github.com/ctuning/ck}{github.com/ctuning/ck} \hspace{1cm} \href{https://cKnowledge.io}{cKnowledge.io}\\
\\
\it{Accepted for \href{https://royalsocietypublishing.org/journal/rsta}{Philosophical Transactions of the Royal Society A}}\\
\\
\it{(Submitted: 12 June 2020)}
}

\date{}

\vskip 0.2in

\maketitle

\begin{abstract}

This article provides the motivation and overview of the \href{https://github.com/ctuning/ck}{Collective Knowledge framework} (CK or cKnowledge).
The CK concept is to decompose research projects into reusable components that encapsulate research artifacts 
and provide unified application programming interfaces (APIs), command-line interfaces (CLIs), meta descriptions and common automation actions for related artifacts.
The CK framework is used to organize and manage research projects as a database of such components.

Inspired by the USB "plug and play" approach for hardware, CK also helps to assemble portable workflows 
that can automatically plug in compatible components from different users and vendors (models, datasets, frameworks, compilers, tools).
Such workflows can build and run algorithms on different platforms and environments 
in a unified way using the customizable CK program pipeline with software detection plugins 
and the automatic installation of missing packages.

This article presents a number of industrial projects in which the modular CK approach was successfully validated 
in order to automate benchmarking, auto-tuning and co-design of efficient software and hardware 
for machine learning (ML) and artificial intelligence (AI) in terms of speed, accuracy, energy, size and various costs.
The CK framework also helped to automate the artifact evaluation process at several computer science conferences
as well as to make it easier to reproduce, compare and reuse research techniques from published papers,
deploy them in production, and automatically adapt them to continuously changing datasets, models 
and systems.

The long-term goal is to accelerate innovation by connecting researchers and practitioners 
to share and reuse all their knowledge, best practices, artifacts, workflows and experimental results
in a common, portable and reproducible format at \href{https://cKnowledge.io}{cKnowledge.io}.

\end{abstract}

\vspace{0.3cm}


\vspace{0.3cm}

]

\section{Motivation}

\begin{figure*}[ht]
  \centering
  \includegraphics[width=1.0\textwidth]{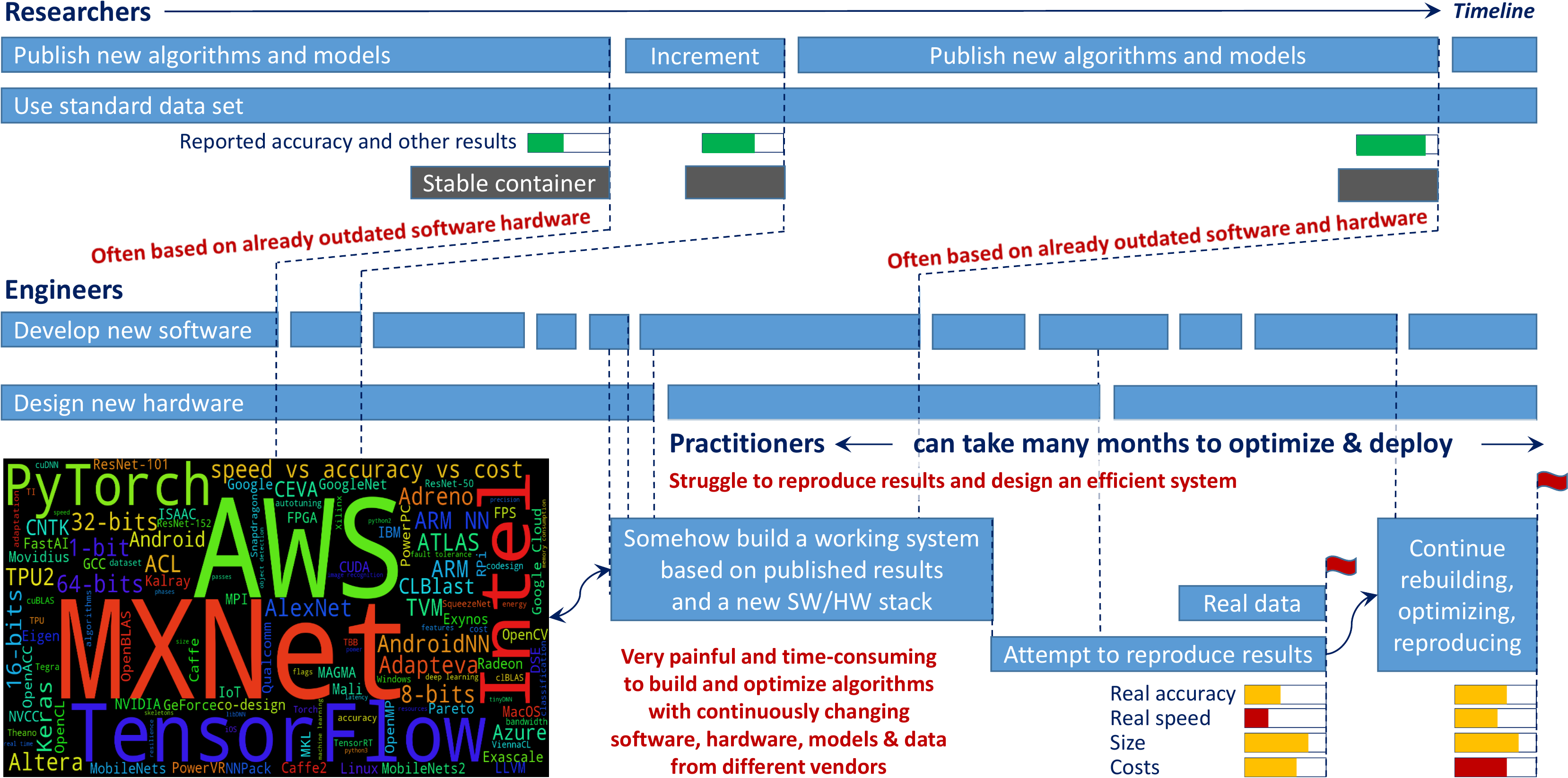}
  \caption{Reproducing research papers and adopting novel techniques in production 
   is a tedious, repetitive and time-consuming process
   because of continuously changing software, hardware, models and datasets, 
   and a lack of common formats and APIs for shared artifacts (code, data, models, experimental results, scripts and so on).}
  \label{fig:timeline1}
\end{figure*}

Ten years ago I developed the \href{https://cTuning.org}{cTuning.org} platform and released all my research code and data 
into the public domain in order to crowdsource the training of our machine learning-based compiler (MILEPOST GCC)~\cite{Fur2009}.
My intention was to accelerate the very time-consuming auto-tuning process and help our compiler 
to learn the most efficient optimizations across real programs, datasets, platforms and environments provided by volunteers. 

We had a positive response from the community and, in just a few days, I collected as many optimization results 
as I had during the entire MILEPOST project.
However, the initial excitement quickly faded when I struggled to reproduce most of the performance numbers and ML model predictions
because even a tiny change in software, hardware, environment or the run-time state of the system could influence performance
while I did not have a mechanism to detect such changes~\cite{mdhs2009,FC2014}.

Worse still, I could not compare these empirical results with other published techniques because they rarely included
the full experiment specification with the shared research code and all the necessary artifacts needed to be able to reproduce results.
Furthermore, it was always very difficult to add new tools, benchmarks and datasets to any research code
because it required numerous changes in different ad-hoc scripts, repetitive recompilation of the whole project 
when new software was released, complex updates of database tables with results and so on~\cite{gfursin_fastpath20}.

These problems motivated me to establish the non-profit cTuning foundation and continue working with the community 
on a common methodology and open-source tools to enable collaborative, reproducible, reusable and trustable R\&D.
We helped to initiate reproducibility initiatives and support artifact
evaluation at several computer science conferences in collaboration with the ACM~\cite{ck-ae,cfkz2016}.
We also promoted sharing of code, artifacts
and results in a unified way along with research papers~\cite{ck-ae-appendix}.

This community service gave me a unique chance to participate in reproducibility studies of more than 100 research papers 
at the International Conference on Architectural Support for Programming Languages and Operating Systems (ASPLOS), 
International Symposium on Code Generation and Optimization (CGO), 
ACM SIGPLAN Annual Symposium on Principles and Practice of Parallel Programming (PPoPP), 
Supercomputing, International Conference on Machine Learning and Systems (MLSys) 
and other computer science conferences over the past five years~\cite{ae-reproduced-papers}.
I also started deploying some of these techniques in production in collaboration with my industrial partners 
in order to better understand all the problems associated with building trustable, reproducible and production-ready computational systems.

This practical experience confirmed my previous findings: while sharing ad-hoc research code, artifacts, trained models,
and Docker images along with research papers is a great step forward, it is only the tip of the reproducibility iceberg~\cite{gfursin_fastpath20}.
The major challenge afterwards is threefold. It is necessary to figure out how to: use research techniques outside original containers
with other data, code and models; run them in a reliable and efficient way 
across rapidly evolving software, heterogeneous hardware and legacy platforms with continuously 
changing interfaces and data formats; and balance multiple characteristics including speed, latency,
accuracy, memory size, power consumption, reliability and costs (Figure~\ref{fig:timeline1}).

\section{Collective Knowledge concept}

When helping to organize the artifact evaluation process at systems and machine learning conferences,
I decided to introduce an Artifact Appendix and a reproducibility checklist~\cite{ck-ae-appendix,ae-reproduced-papers}.
My goal was to help researchers to describe how to configure, build, run, validate and compare research techniques
in a unified way across different conferences and journals.
It also led me to notice that most research projects use some ad-hoc scripts, often with hardwired paths, 
to perform the same repetitive tasks including downloading models and datasets, detecting platform properties,
installing software dependencies, building research code, running experiments, validating outputs, 
reproducing results, plotting graphs and generating papers~\cite{gfursin_fastpath20}.

This experience motivated me to search for a solution for automating such common tasks and making them reusable
and customizable across different research projects, platforms and environments. 
First, I started looking at related tools that were introduced to automate experiments
and make research more reproducible:

\begin{itemize} 

\item
 Workflow frameworks such as MLFlow~\cite{Zaharia2018AcceleratingTM}, 
 Kedro~\cite{kedro}, Amazon SageMaker~\cite{sagemaker},
 Kubeflow~\cite{kubeflow}, Apache Taverna~\cite{10.1093/nar/gkt328},
 popper~\cite{7965226}, CommonWL~\cite{Amstutz2016}
 and many others help to abstract and automate data science operations. 
 They are very useful for data scientists but do not yet provide 
 a universal mechanism to automatically build and run algorithms across different platforms, 
 environments, libraries, tools, models and datasets.
 Researchers and engineers often have to implement this functionality
 for each new project from scratch, which can become very complicated 
 particularly when targeting new hardware, embedded devices, TinyML and IoT.
 
\item
 Machine learning benchmarking initiatives such as MLPerf~\cite{reddi2019mlperf}, 
 MLModelScope~\cite{li2019acrossstack} and Deep500~\cite{deep500}
 attempt to standardise machine learning model benchmarking
 and make it more reproducible.
 However, production deployment, integration with complex systems 
 and adaptation to continuously changing user environments, platforms, tools,
 and data are currently out of their scope.
 
\item
 Package managers such as Spack~\cite{spack} and EasyBuild~\cite{easybuild}
 are very useful for rebuilding and fixing the whole software environment.
 However, the integration with workflow frameworks and automatic adaptation 
 to existing environments, native cross-compilation
 particularly for embedded devices, and support for non-software packages 
 (models, datasets, scripts) are still in progress.

\item
 Container technology such as Docker~\cite{10.5555/2600239.2600241} is very useful
 for preparing and sharing stable software releases. 
 However, it hides the software chaos rather than solving it, lacks common APIs
 for research projects, requires enormous amount of space, has very poor support for embedded
 devices, and does not yet help integrate models with existing projects,
 legacy systems and user data.

\item
 PapersWithCode platform~\cite{paperswithcode} helps to find relevant research code
 for machine learning papers and keeps track of state-of-the-art 
 ML research using public scoreboards with non-validated experimental results
 from papers. 
 However, my experience reproducing research papers suggests that sharing ad-hoc research code 
 is not enough to make research techniques reproducible, customizable,
 portable and trustable~\cite{gfursin_fastpath20}.
  
\end{itemize} 

While testing all these useful tools and analyzing the Jupyter notebooks, Docker images and GitHub repositories 
shared alongside research papers, I started thinking that it would be possible to reorganize them as some sort of database
of reusable components with a common API, command line, web interface and meta description.
We could then reuse artifacts and some common automation actions across different projects 
while applying DevOps principles to research projects.

Furthermore, we could also gradually abstract and interconnect all existing tools rather than rewriting or substituting them.
This, in turn, helped to create "plug and play" workflows that could automatically connect compatible components 
from different users and vendors (models, datasets, frameworks, compilers, tools, platforms)
while minimising manual interventions and providing a common interface for all shared research projects.

I called my project Collective Knowledge (CK) because my goal was to connect researchers and practitioners
in order to share their knowledge, best practices and research results in the form of reusable components,
portable workflows, and automation actions with common APIs and meta descriptions.

\section{CK framework and an open CK format}

\begin{figure*}[ht]
  \centering
  \includegraphics[width=1.0\textwidth]{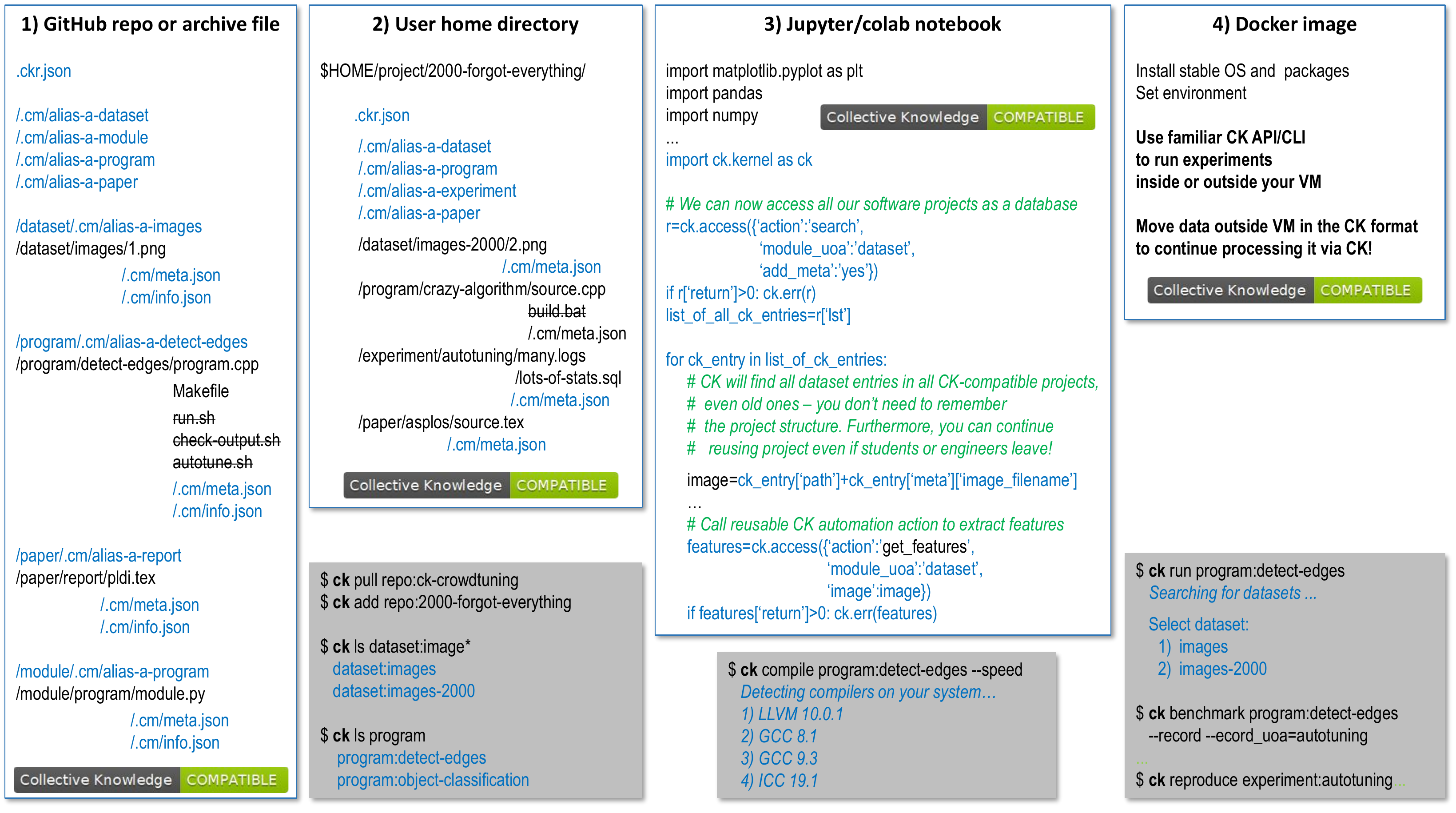}
  \caption{Organizing software projects as a human-readable database of reusable components
  that wrap artifacts and provide unified APIs, CLI, meta descriptions 
  and common automation actions for related artifacts. 
  The CK APIs and control files are highlighted in blue.}
  \label{fig:ck-concept}
\end{figure*}

I have developed the CK framework as a small Python library with minimal dependencies
to be very portable while offering the possibility of being re-implemented in other languages
such as C, C++, Java and Go.
The CK framework has a unified CLI, a Python API
and a JSON-based web service to manage \emph{CK repositories}
and add, find, update, delete, rename and move \emph{CK components} 
(sometimes called \emph{CK entries} or \emph{CK data})~\cite{ck3}.

CK repositories are human-readable databases of reusable CK components that can be created 
in any local directory and inside containers, pulled from GitHub and similar services, 
and shared as standard archive files~\cite{ck-repos}.
\emph{CK components} simply wrap (encapsulate) user artifacts and provide an extensible JSON meta description
with \emph{common automation actions}~\cite{ck-actions} for related artifacts.

Automation actions are implemented using \emph{CK modules}, which are Python modules
with functions exposed in a unified way via CK API and CLI 
and which use dictionaries for input and output (extensible and unified CK input/output (I/O)).
The use of dictionaries makes it easier to support continuous integration tools
and web services and to extend the functionality while keeping backward compatibility.
The unified I/O also makes it possible to reuse such actions across projects
and to chain them together into unified CK pipelines and workflows.

As I wanted CK to be non-intrusive and technology neutral, I decided to use 
a simple 2-level directory structure to wrap user artifacts into CK components
as shown in Figure~\ref{fig:ck-concept}. 
The root directory of the CK repository contains the \emph{.ckr.json} file to describe the repository 
and specify dependencies on other CK repositories to explicitly reuse their components and automation actions. 

CK uses \emph{.cm} directories (Collective Meta) similar to \emph{.git} to store meta information of all components 
as well as Unique IDs to be able to find components even 
if their user-friendly names have changed over time (\emph{CK alias}).
CK modules are always stored in \emph{module / \textless CK module name \textgreater} directories in the CK repository.
CK components are stored in \emph{\textless CK module name \textgreater / \textless CK data name \textgreater} directories.
Each \emph{CK component} has a \emph{.cm} directory with the \emph{meta.json} file 
describing a given artifact and \emph{info.json} file to keep the provenance 
of a given artifact including copyrights, licenses, creation date, 
names of all contributors, and so on.

The CK framework has an internal \emph{default CK repository}
with \emph{stable CK modules} and the most commonly used automation actions
across many research projects.
When the CK framework is used for the first time, it also creates a \emph{local CK repository}
in the user space to be used as a scratch pad.

After discussing the CK CLI with colleagues, I decided to implement it in a similar 
way to natural language to make it easier for users to remember the commands:

\begin{strip}
\begin{lstlisting}[language=bash, basicstyle=\footnotesize]
ck <action> <CK module name> (flags) (@input.json or @input.yaml)
ck <action> <CK module name>:<CK data name> (flags)
ck <action> <CK repository name>:<CK module name>:<CK data name>
\end{lstlisting}
\end{strip}

The next example demonstrates how to compile and run the shared automotive benchmark
on any platform, and then create a copy of the CK \emph{program component}:

\clearpage
\begin{strip}
\vbox{
\begin{lstlisting}[language=bash, basicstyle=\footnotesize]
pip install ck

ck pull repo --url=https://github.com/ctuning/ck-crowdtuning

ck search dataset --tags=jpeg

ck search program:cbench-automotive-*

ck find program:cbench-automotive-susan

ck load program:cbench-automotive-susan

ck help program

ck compile program:cbench-automotive-susan --speed
ck run program:cbench-automotive-susan --env.OMP_NUM_THREADS=4

ck run program --help

ck cp program:cbench-automotive-susan local:program:new-program-workflow
ck find program:new-program-workflow

ck benchmark program:new-program-workflow --record --record_uoa=my-test
ck replay experiment:my-test
\end{lstlisting}
}
\end{strip}

The CK program module describes dependencies on software detection plugins 
and meta packages using simple tags and version ranges that the community
has to agree on:

\begin{strip}
\vbox{
\begin{lstlisting}[basicstyle=\footnotesize]
 {
   "compiler": {
     "name": "C++ compiler",
     "sort": 10,
     "tags": "compiler,lang-cpp"
   },
   "library": {
     "name": "TensorFlow C++ API",
     "sort": 20,
     "version_from": [1,13,1],
     "version_to": [2,0,0],
     "no_tags": "tensorflow-lite",
     "tags": "lib,tensorflow,vstatic"
   }
 }
\end{lstlisting}
}
\end{strip}

I also implemented a simple \emph{access} function in the CK Python API 
to access all the CK functionality in a very simple and unified way:

\begin{strip}
\vbox{
\begin{lstlisting}[language=python, basicstyle=\footnotesize]
import ck.kernel as ck

# Equivalent of "ck compile program:cbench-automotive-susan --speed"
r=ck.access({'action':'compile', 
             'module_uoa':'program', 'data_uoa':'cbench-automotive-susan', 
             'speed':'yes'})
if r['return']>0: return r # unified error handling 

print (r)

# Equivalent of "ck run program:cbench-automotive-susan --env.OMP_NUM_THREADS=4
r=ck.access({'action':'run', 
             'module_uoa':'program', 'data_uoa':'cbench-automotive-susan', 
             'env':{'OMP_NUM_THREADS':4}})
if r['return']>0: return r # unified error handling 

print (r)
\end{lstlisting}
}
\end{strip}

Such an approach allowed me to connect colleagues, students, researchers and engineers
from different workgroups to implement, share and reuse automation actions and CK components 
rather than reimplementing them from scratch.
Furthermore, the Collective Knowledge concept supported the so-called FAIR principles 
to make data findable, accessible, interoperable and reusable~\cite{wilkinson2016fair} 
while also extending it to code and other research artifacts.


\section{CK components and workflows to automate machine learning and systems research}

One of the biggest challenges I have faced throughout my research career automating the co-design process
of efficient and self-optimizing computing systems has been how to deal with rapidly evolving hardware, models, datasets, compilers and research techniques.
It is for this reason that my first use case for the CK framework was to work with colleagues
to collaboratively solve these problems and thus enable trustable, reliable and efficient computational systems 
that can be easily deployed and used in the real world.

We started using CK as a flexible playground to decompose complex computational systems
into reusable, customizable and non-virtualized CK components 
while agreeing on their APIs and meta descriptions.
As the first step, I implemented basic actions that could automate
the most common R\&D tasks that I encountered during artifact evaluation
and in my own research on systems and machine learning~\cite{gfursin_fastpath20}.
I then shared the automation actions to analyze platforms and user environments
in a unified way, detect already installed code, data and ML models
(\emph{CK software detection plugins}~\cite{ck-soft-plugins}),
and automatically download, install and cross-compile missing packages
(\emph{CK meta packages}~\cite{ck-meta-packages}).
At the same time, I provided initial support for different compilers, operating systems (Linux, Windows, MacOS,
Android) and hardware from different vendors including Intel, Nvidia, Arm, Xilinx, AMD, Qualcomm, Apple, Samsung and Google.

Such an approach allowed my collaborators~\cite{ck-projects} to create, share and reuse 
different CK components with unified APIs to detect, install and use 
different AI and ML frameworks, libraries, tools, compilers, models and datasets.
The unified automation actions, APIs and JSON meta descriptions of all CK components
also helped us to connect them into platform-agnostic, portable and customizable program pipelines 
(workflows) with a common interface across all research projects while applying the DevOps methodology.

Such "plug\&play" workflows~\cite{cm:29db2248aba45e59:c4b24bff57f4ad07} can automatically adapt 
to evolving environments, models, datasets and non-virtualized platforms by automatically detecting the properties 
of a target platform, finding all required dependencies and artifacts (code, data and models) 
on a user platform with the help of CK software detection plugins~\cite{ck-soft-plugins}, 
installing missing dependencies using portable CK meta packages~\cite{ck-meta-packages}, 
building and running code, and unifying and testing outputs~\cite{ck-portable-workflows}.
Moreover, rather than substituting or competing with existing tools, the CK approach
helped to abstract and interconnect them in a relatively simple and non-intrusive way.

CK also helps to protect user workflows whenever external files or packages 
are broken, disappear or move to another URL because it is possible to fix such 
issues in a shared CK meta package without changing existing workflows.
For example, our users have already taken advantage of this functionality when the Eigen library 
moved from BitBucket to GitLab and when the old ImageNet dataset was no longer supported
but could still be downloaded via BitTorrent and other peer-to-peer services.

\begin{figure*}[htb]
  \centering
  \includegraphics[width=1.0\textwidth]{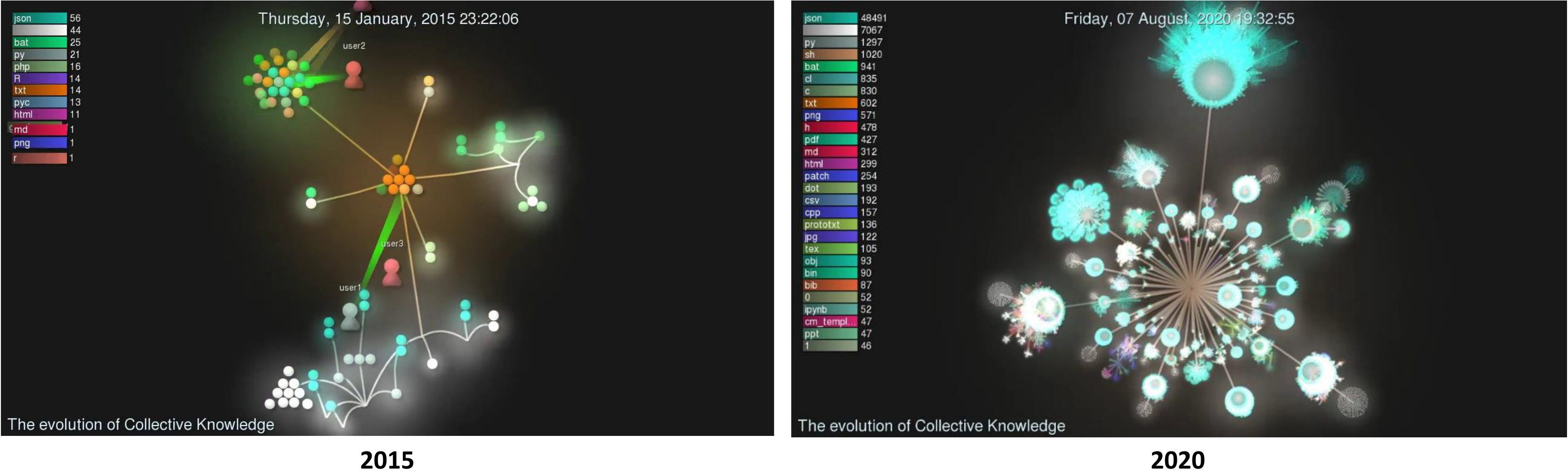}
  \caption{Collective Knowledge framework provided a flexible playground 
   for researchers and practitioners to decompose 
   complex computational systems into reusable components while agreeing
   on APIs and meta descriptions. 
   Over the past 5 years, the Collective Knowledge project 
   has grown to thousands of reusable CK components and automation actions.
  }
  \label{fig:ck-evolution}
\end{figure*}

Modular CK workflows can help to keep track of the information flow within such workflows, 
gradually expose configuration and optimization parameters as vectors via dictionary-based I/O,
and combine public and private code and data.
They also help to monitor, model and auto-tune system behaviour, 
retarget research code and machine learning models 
to different platforms from data centers to edge devices, integrate them with legacy systems, 
and reproduce results.

Furthermore, we can use CK workflows inside standard containers such as Docker 
while providing a unified CK API to customize, rebuild and adapt them
to any platform (\emph{Adaptive CK container})~\cite{ck-adaptive-containers}
thus making research techniques truly portable, reproducible and reusable.
I envision that such adaptive CK containers and portable workflows can complement existing 
marketplaces to deliver portable, customizable, trustable and efficient AI and ML solutions
that can be continuously optimized across diverse models, datasets, frameworks,
libraries, compilers and run-time systems.

In spite of some initial doubts, my collaborative CK approach has worked well to decompose complex research projects and computational systems into reusable components while agreeing on common APIs and meta descriptions: the CK functionality evolved 
from just a few core CK modules and abstractions to hundreds of CK modules~\cite{ck-modules} 
and thousands of reusable CK components and workflows~\cite{ck-platform}
to automate the most repetitive research tasks particularly for AI, ML and systems R\&D as shown in Figure~\ref{fig:ck-evolution}.

\section{CK use cases}

\subsection{Unifying benchmarking, auto-tuning and machine learning}

The MILEPOST and cTuning projects were like the Apollo mission: 
on the one hand, we managed to demonstrate that it was indeed possible to crowdsource
auto-tuning and machine learning across multiple users
to automatically co-design efficient software and hardware~\cite{Fur2009,29db2248aba45e59:a31e374796869125}.
On the other hand, we exposed so many issues when using machine learning for system optimization in the real world
that I had to stop this research and have since focused on solving a large number of related engineering problems.

For this reason, my first step was to test the CK concept by converting all artifacts and automation actions
from all my past research projects related to self-optimizing computer systems into reusable CK components
and workflows.
I shared them with the community in the CK-compatible Git repositories~\cite{ck-repos}
and started reproducing experiments from my own or related research projects~\cite{reproduced-results}.

\begin{figure*}[htb]
  \centering
  \includegraphics[width=1.0\textwidth]{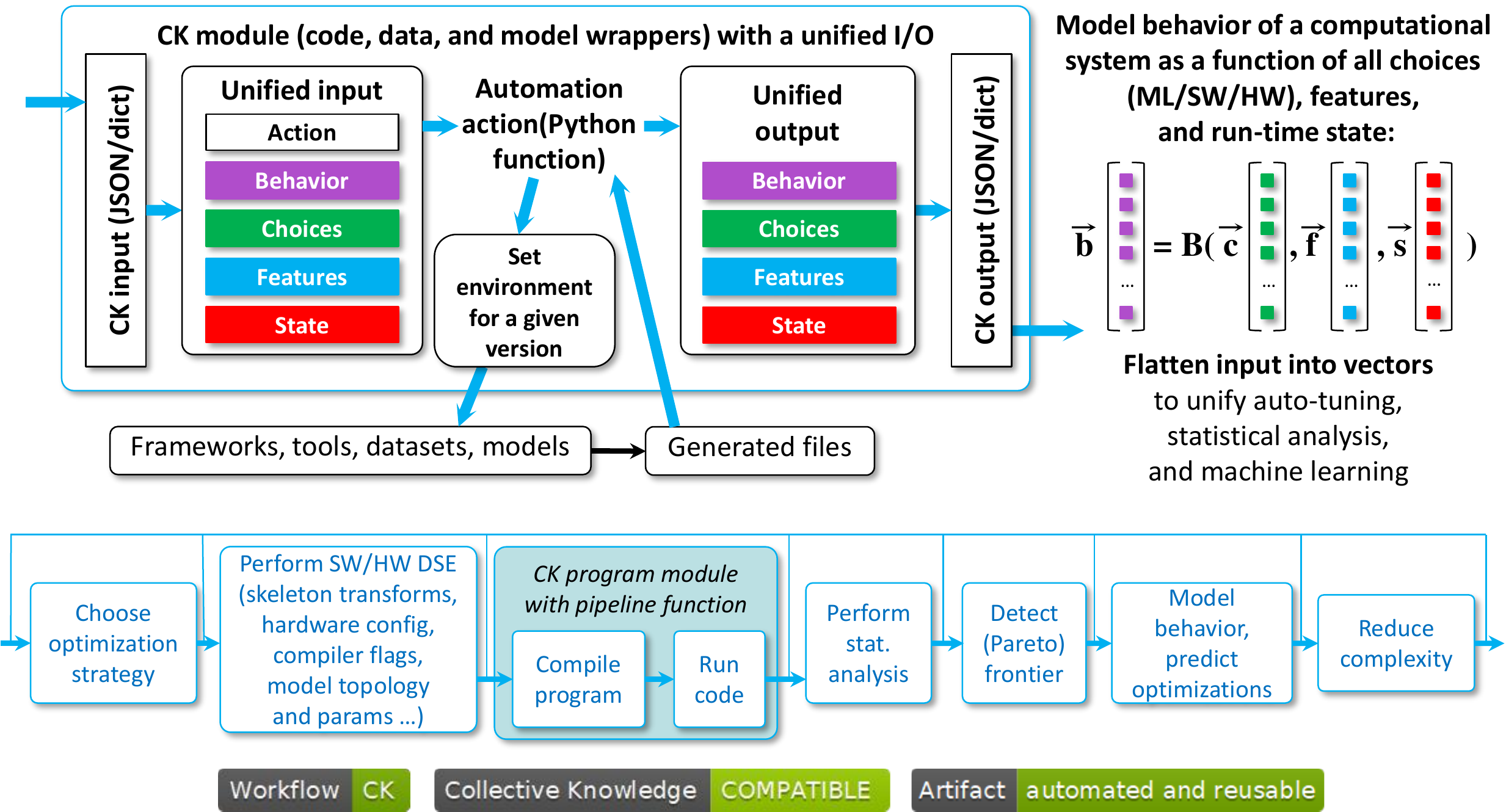}
  \caption{Portable, customizable and reusable program pipeline (workflow) assembled from the CK components 
  to unify benchmarking, auto-tuning and machine learning. It is possible to gradually expose
  design search space, optimizations, features and the run-time state from all CK components to make it easier
  to unify performance modelling and auto-tuning.}
  \label{fig:ck-pipeline}
\end{figure*}

I then implemented a customizable and portable program pipeline as a CK module to unify benchmarking and auto-tuning
while supporting all research techniques and experiments from my PhD research and the MILEPOST project~\cite{reproduce-milepost}.
Such a pipeline could perform compiler auto-tuning and software/hardware co-design 
combined with machine learning in a unified way across different programs, datasets, frameworks,
compilers, ML models and platforms as shown in Figure~\ref{fig:ck-pipeline}.

The CK program pipeline helps users to gradually expose different design choices and optimization parameters 
from all CK components (models, frameworks, compilers, run-time systems, hardware) via unified CK APIs and meta descriptions
and thus enable the whole ML and system auto-tuning.
It also helps users to keep track of all information passed between components in complex
computational systems to ensure the reproducibility of results while finding the most efficient 
configuration on a Pareto frontier in terms of speed, accuracy, energy and other characteristics 
also exposed via unified CK APIs. 
More importantly, it can be now reused and extended in other real-world projects~\cite{ck-projects}.

\subsection{Bridging the growing gap between education, research and practice}

During the MILEPOST project I noticed how difficult it is to start using research techniques
in the real world.
New software, hardware, datasets and models are usually available at the end of such 
research projects making it very challenging, time-consuming and costly to make research software 
work with the latest systems or legacy technology.

That prompted us to organize a proof-of-concept project with the Raspberry Pi foundation
to check if it was possible to use portable CK workflows and components
to enable sustainable research software that can automatically adapt to 
rapidly evolving systems.
We also wanted to provide an open-source tool for students and researchers to help them share
their code, data and models as reusable, portable and customizable workflows and artifacts
- something now known as FAIR principles~\cite{wilkinson2016fair}.

For this purpose, we decided to reuse the CK program workflow to demonstrate 
that it was possible to crowdsource compiler auto-tuning across any Raspberry Pi device, 
with any environment and any version of any compiler (GCC or LLVM)
to automatically improve the performance and code size of the most popular RPi applications.
CK helped to automate experiments, collect performance numbers on live CK scoreboards 
and automatically plug in CK components with various machine learning and predictive analytics techniques 
including decision trees, nearest neighbour classifiers, support vector machines (SVM) and deep learning,
to automatically learn the most efficient optimizations~\cite{cm:29db2248aba45e59:c4b24bff57f4ad07}.

With this project, we demonstrated that it was possible to reuse portable CK workflows 
and let users participate in collaborative auto-tuning (crowd-tuning) on new systems
while sharing best optimizations and unexpected behaviour on public CK scoreboards, 
even after the project.

\subsection{Co-designing efficient software and hardware for AI, ML and other emerging workloads}

\begin{figure*}[ht]
  \centering
  \includegraphics[width=1.0\textwidth]{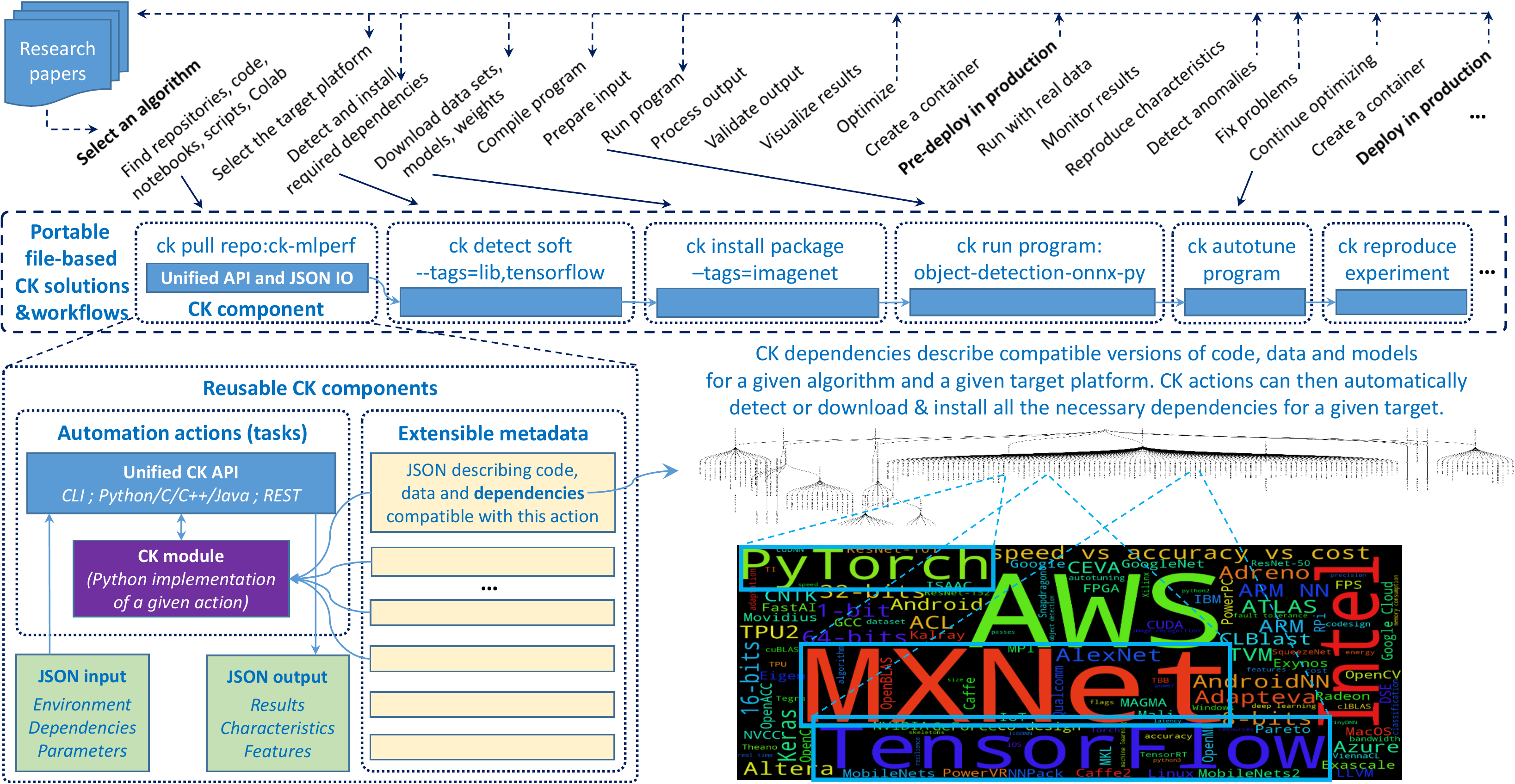}
  \caption{The use of the CK framework to automate benchmarking, optimization and co-design
   of efficient software and hardware for machine learning and artificial intelligence.
   The goal is to make it easier to reproduce, reuse, adopt and build upon ML and systems research.}
  \label{fig:ck-ml}
\end{figure*}

While helping companies to assemble efficient software and hardware for image classification, object detection 
and other emerging AI and ML workloads, I noticed that it can easily take several months 
to build an efficient and reliable system before moving it to production.

This process is so long and tedious because one has to navigate a multitude of design decisions 
when selecting components from different vendors for different applications
(image classification, object detection, natural language processing (NLP), speech recognition and many others)
while trading off speed, latency, accuracy, energy and other costs: 
what network architecture to deploy and how to customize it (ResNet, MobileNet, GoogleNet, SqueezeNet, SSD, GNMT); 
what framework to use (PyTorch vs. MXNet vs. TensorFlow vs. TF Lite vs. Caffe vs. CNTK); 
what compilers to use (XLA vs. nGraph vs. Glow vs. TVM);
what libraries and which optimizations to employ (ArmNN vs. MKL vs. OpenBLAS vs. cuDNN). 
This is generally a consequence of the target hardware platform 
(CPU vs. GPU vs. DSP vs. FPGA vs. TPU vs. Edge TPU vs. numerous accelerators).
Worse still, this semi-manual process is usually repeated from scratch for each new version 
of hardware, models, frameworks, libraries and datasets.

My modular CK program pipeline shown in Figure~\ref{fig:ck-ml} helped to automate this process.
We extended it slightly to plug in different AI and ML algorithms, datasets, models, frameworks and libraries
for different hardware such as CPU, GPU, DSP and TPU and different target platforms
from servers to Android devices and IoT~\cite{ck-projects}.
We also customized this ML workflow with the new CK plugins 
that performed pre- and post-processing of different models and datasets 
to make them compatible with different frameworks, backends and hardware
while unifying benchmarking results such as throughput, latency,
mAP (mean Average Precision), recall and other characteristics.
We also exposed different design and optimization parameters including model topology, batch sizes,
hardware frequency, compiler flags and so on.

Eventually, CK allowed us to automate and systematise design space exploration (DSE)
of AI/ML/software/hardware stacks and distribute it across diverse platforms and environments~\cite{ck-dse-example}.
This is possible because CK automatically detects all necessary dependencies on any platform, 
installs and/or rebuilds the prerequisites, runs experiments, and records all results together with the complete
experiment configuration (resolved dependencies and their versions,
environment variables, optimization parameters and so on)
in a unified JSON format inside CK repositories.
CK also ensured the reproducibility of results while making it easier to analyze and visualize 
results locally using Jupyter notebooks and standard toolsets or within workgroups 
using universal CK dashboards also implemented as CK modules~\cite{reproduced-results}.

Note that it is also possible to share the entire experimental setup in the CK format 
inside Docker containers, thus automating all the DSE steps using the unified CK API 
instead of trying to figure them out from the ReadMe files.
This method enables CK-powered adaptive containers that help users to start using and customizing 
research techniques across diverse software and hardware from servers to mobile devices
in just a few simple steps while sharing experimental results within workgroups or along research papers in the CK format, 
reproducing and comparing experiments, and even automatically reporting unexpected behaviour such as bugs and mispredictions~\cite{unexpected-behavior}.

Eventually, I managed to substitute my original cTuning framework completely
with the modular, portable, customizable and reproducible experimental framework
while addressing most of the engineering and reproducibility issues 
exposed by the MILEPOST and cTuning projects~\cite{Fur2009}.
It also helped me to return to my original research on lifelong benchmarking, optimization and co-design
of efficient software and hardware for emerging workloads, including machine learning
and artificial intelligence.

\subsection{Automating MLPerf and enabling portable MLOps}

The modularity of my portable CK program workflow helped to enable portable MLOps when combined with AI and ML components, FAIR principles and the DevOps methodology.
For example, my CK workflows and components were reused and extended by General Motors and dividiti
to collaboratively benchmark and optimize deep learning implementations across diverse devices 
from servers to the edge~\cite{GM}.
They were also used by Amazon to enable scaling of deep learning on AWS using C5 instances 
with MXNet, TensorFlow and BigDL~\cite{ck-amazon}.
Finally, the CK framework made it easier to prepare, submit and reproduce MLPerf inference benchmark 
results ("fair and useful benchmark for measuring training and inference performance of ML hardware, software, and services")~\cite{reddi2019mlperf,mlperf-inference-results}.
These results are aggregated in the public optimization repository~\cite{reproduced-results}
to help users find and compare different AI/ML/software/hardware stacks in terms of speed, accuracy, power consumption, costs and other collected metrics.

\subsection{Enabling reproducible papers with portable workflows and reusable artifacts}

\begin{figure*}[ht]
  \centering
  \includegraphics[width=1.0\textwidth]{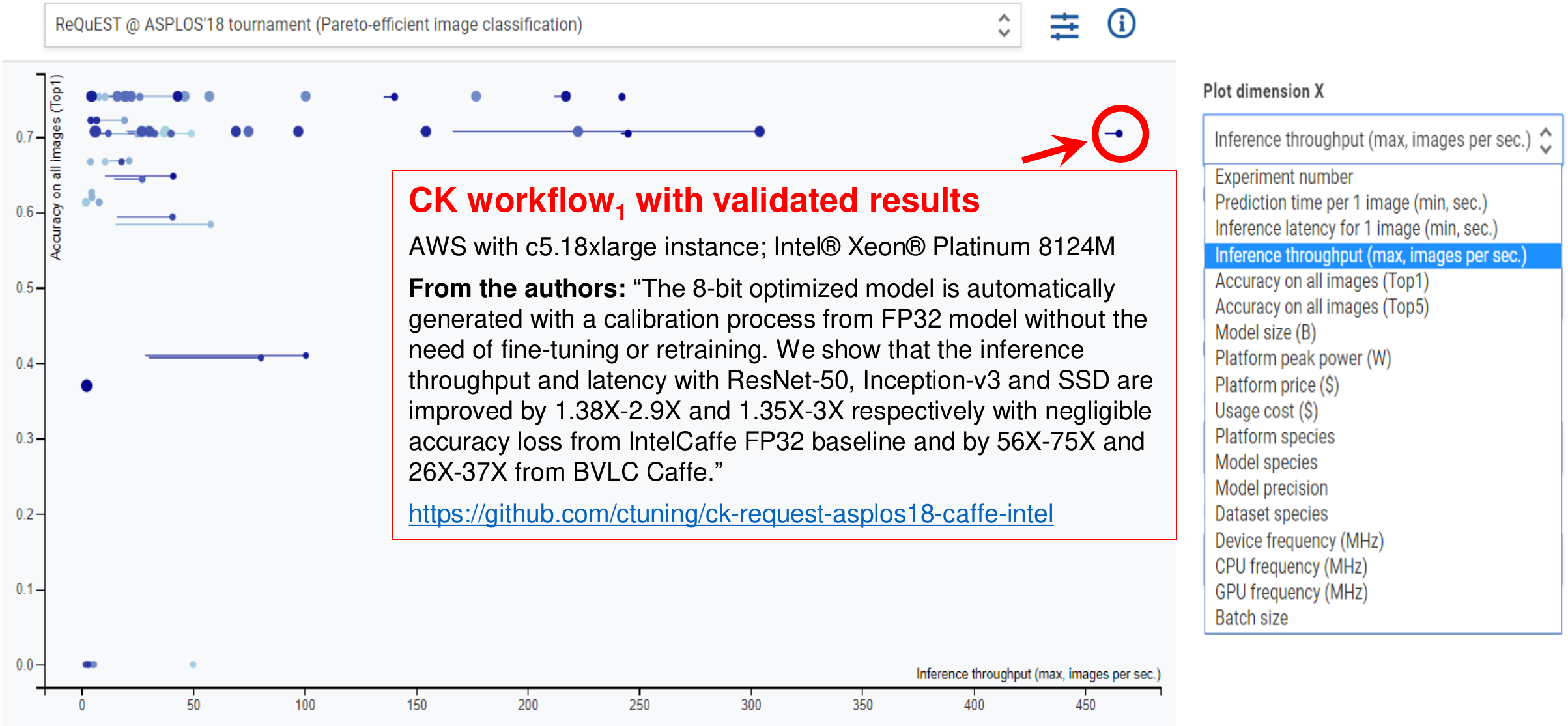}
  \caption{The CK dashboard with reproduced results from the ACM ASPLOS-REQUEST'18 tournament to co-design Pareto-efficient image classification in terms of speed, accuracy, energy and other costs. Each paper submission was accompanied by the portable CK workflow to be able to reproduce and compare results in a unified way.}
  \label{fig:ck-request}
\end{figure*}

Ever since my very first research project, I wanted to be able to easily
find all artifacts (code, data, models) from research papers, reproduce
and compare results in just a few clicks, and immediately test research
techniques in the real world with different platforms, environments and
data.
It is for that reason that one of my main goals when designing the CK framework was to use portable CK workflows for these purposes.

I had the opportunity to test my CK approach when co-organizing the first reproducible tournament 
at the ACM ASPLOS'18 conference (the International Conference on Architectural Support for Programming Languages and Operating Systems). The tournament required participants 
to co-design Pareto-efficient systems for deep learning in terms of speed, accuracy, energy,
costs and other metrics~\cite{REQ}.
We wanted to extend existing optimization competitions, tournaments and hackathons including Kaggle~\cite{kaggle},
ImageNet~\cite{imagenet-challenge}, the Low-Power Image Recognition Challenge (LPIRC)~\cite{lpirc},
DAWNBench (an end-to-end deep learning benchmark and competition)~\cite{DAWNBench},
and MLPerf~\cite{reddi2019mlperf,MLP} with a customizable experimental 
framework for collaborative and reproducible optimization
of Pareto-efficient software and hardware stack for deep learning
and other emerging workloads.

This tournament helped to validate my CK approach for reproducible papers.
The community submitted five complete implementations (code, data, scripts, etc.)
for the popular ImageNet object classification challenge. 
We then collaborated with the authors to convert their artifacts into
the CK format, evaluate the converted artifacts on the original or similar platforms,
and reproduce the results based on the rigorous artifact evaluation methodology~\cite{ck-ae}.
The evaluation metrics included accuracy on the ImageNet validation set (50,000 images), latency (seconds per image),
throughput (images per second), platform price (dollars) and peak power consumption (Watts).
Since collapsing all metrics into one to select a single winner often results in over-engineered solutions, 
we decided to aggregate all reproduced results on a universal CK scoreboard shown in Figure~\ref{fig:ck-request}
and let users select multiple implementations from a Pareto frontier, based on their requirements
and constraints.

We then published all five papers with our unified artifact appendix~\cite{ck-ae-appendix} and a set of ACM reproducibility badges
in the ACM Digital Library~\cite{10.1145/3229762}, accompanied by adaptive CK containers (CK-powered Docker) and portable CK workflows
covering a very diverse model/software/hardware stack:

\begin{itemize}

 \item \textbf{Models:} MobileNets, ResNet-18, ResNet-50, Inception-v3, VGG16, AlexNet, SSD.
 \item \textbf{Data types:} 8-bit integer, 16-bit floating-point (half), 32-bit floating-point (float).
 \item \textbf{AI frameworks and libraries:} MXNet, TensorFlow, Caffe, Keras, Arm Compute Library, cuDNN, TVM, NNVM.
 \item \textbf{Platforms:} Xilinx Pynq-Z1 FPGA, Arm Cortex CPUs and Arm Mali GPGPUs (Linaro HiKey960 and T-Firefly RK3399), a farm of Raspberry Pi devices, NVIDIA Jetson TX1 and TX2, and Intel Xeon servers in Amazon Web Services, Google Cloud and Microsoft Azure.

\end{itemize}

The reproduced results also exhibited great diversity:

\begin{itemize}

 \item \textbf{Latency:} 4 .. 500 milliseconds per image
 \item \textbf{Throughput:} 2 .. 465 images per second
 \item \textbf{Top 1 accuracy:} 41 .. 75 percent
 \item \textbf{Top 5 accuracy:} 65 .. 93 percent
 \item \textbf{Model size (pre-trained weights):} 2 .. 130 megabytes
 \item \textbf{Peak power consumption:} 2.5 .. 180 Watts
 \item \textbf{Device frequency:} 100 .. 2600 megahertz
 \item \textbf{Device cost:} 40 .. 1200 dollars
 \item \textbf{Cloud usage cost:} 2.6E-6 .. 9.5E-6 dollars per inference

\end{itemize}

The community can now access all the above CK workflows under permissive licenses and continue
collaborating on them via dedicated GitHub projects with CK repositories.
These workflows can be automatically adapted to new platforms and environments by either detecting already
installed dependencies (frameworks, libraries, datasets) or rebuilding dependencies 
using CK meta packages supporting Linux, Windows, MacOS and Android. 
They can be also extended to expose new design and optimization choices such as quantization, 
as well as evaluation metrics such as power or memory consumption.
We also used these CK workflows to crowdsource the design space exploration 
across devices provided by volunteers such as mobile phones, laptops and servers
with the best solutions aggregated on live CK scoreboards~\cite{reproduced-results}.

\begin{figure*}[ht]
  \centering
  \includegraphics[width=1.0\textwidth]{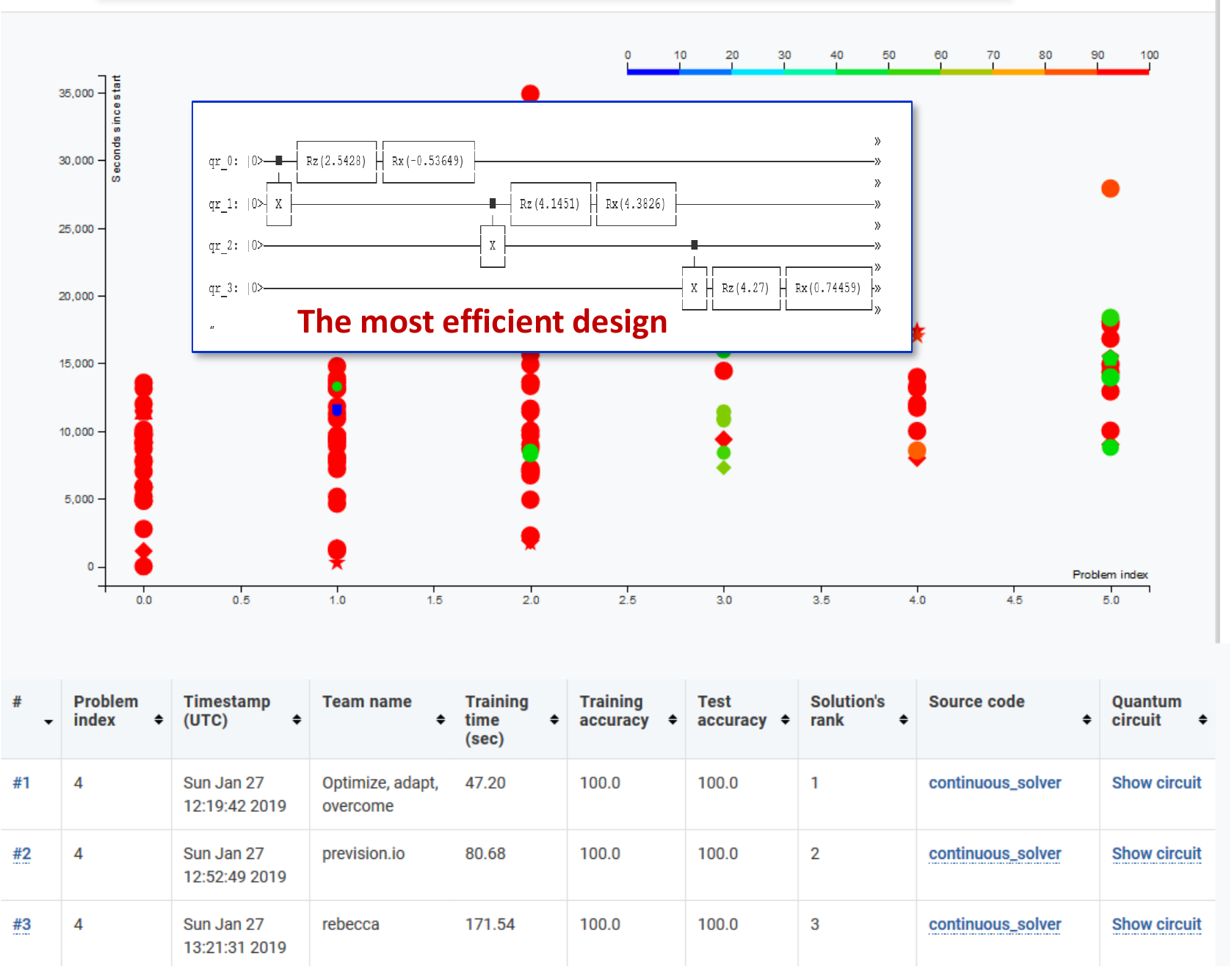}
  \caption{The CK dashboard connected with portable CK workflows to visualize and compare public results from reproducible Quantum Hackathons.
Over 80 participants worked together to solve a quantum machine learning problem and minimise time to solution.
}
  \label{fig:ck-quantum}
\end{figure*}

After validating that my portable CK program workflow can support 
reproducible papers for deep learning systems, I decided to conduct
one more test to check if CK could also support quantum computing R\&D.
Quantum computers have the potential to solve certain problems
dramatically faster than conventional computers, with applications
in areas such as machine learning, drug discovery, materials,
optimization, finance and cryptography.
However it is not yet known when the first demonstration of quantum advantage will be achieved, or what shape it will take.

This led me to co-organize several Quantum hackathons similar to the REQUEST tournament~\cite{QCK} with IBM, Rigetti, Riverlane and dividiti.
My main goal was to check if we could aggregate and share multidisciplinary knowledge about the state-of-the-art in quantum computing
using portable CK workflows that can run on classical hardware and quantum platforms from IBM, Rigetti and other companies, 
can be connected to a public dashboard to simplify reproducibility and comparison of different algorithms across different platforms,
and can be extended by the community even after hackathons.

Figure~\ref{fig:ck-quantum} shows the results from one of such Quantum hackathons 
where over 80 participants, from undergraduate and graduate students 
to startup founders and experienced professionals 
from IBM and CERN, worked together 
to solve a quantum machine learning problem designed by Riverlane.
All participants were given some labelled quantum data and asked to develop algorithms for solving a classification problem.

We also taught participants how to perform these experiments in a collaborative, reproducible and automated
way using the CK framework so that the results could be transfered to industry.
For example, we introduced the CK repository with workflows and components
for the Quantum Information Science Kit (QISKit) - an open source software development kit (SDK) 
for working IBM Q quantum processors~\cite{ck-qiskit}.
Using the CK program workflow from this repository, the participants were able to start running quantum experiments with a standard CK command:

\begin{lstlisting}[language=bash, basicstyle=\footnotesize]
ck pull repo --url=https://github.com/ctuning/ck-qiskit
ck run program:qiskit-demo --cmd_key=quantum_coin_flip
\end{lstlisting}

Whenever ready, the participants could submit their solutions to the public CK dashboards
to let other users validate and reuse their results~\cite{reproduced-results,QCK}.

Following the successful validation of portable CK workflows for reproducible papers,
I continued collaborating with ACM~\cite{ck-acm} 
and ML and systems conferences to automate the tedious artifact evaluation process~\cite{ck-ae,gfursin_fastpath20}.
For example, we developed several CK workflows to support 
the Student Cluster Competition Reproducibility Challenge (SCC)
at the Supercomputing conference~\cite{ck-scc}.
We demonstrated that it was possible to reuse the CK program workflow
to automate the installation, execution, customization and validation 
of the SeisSol application (Extreme Scale Multi-Physics Simulations 
of the Tsunamigenic 2004 Sumatra Megathrust Earthquake)~\cite{10.1145/3126908.3126948}
from the SC18 Student Cluster Competition Reproducibility Challenge
across several supercomputers and HPC clusters~\cite{ck-scc18}.
We also showed that it was possible to abstract HPC job managers 
including Slurm and Flux and connect them with our portable CK workflows.

Some authors already started using CK to share their research research artifacts
and workflows at different ML and systems conferences 
during artifact evaluation~\cite{reproduced-papers-with-ck-workflows}.
My current goal is to make the CK onboarding as simple as possible
and help researchers to automatically convert their ad-hoc artifacts 
and scripts into CK workflows, reusable artifacts, adaptive containers
and live dashboards.

\subsection{Connecting researchers and practitioners to co-design efficient computational systems}

\begin{figure*}[ht]
  \centering
  \includegraphics[width=1.0\textwidth]{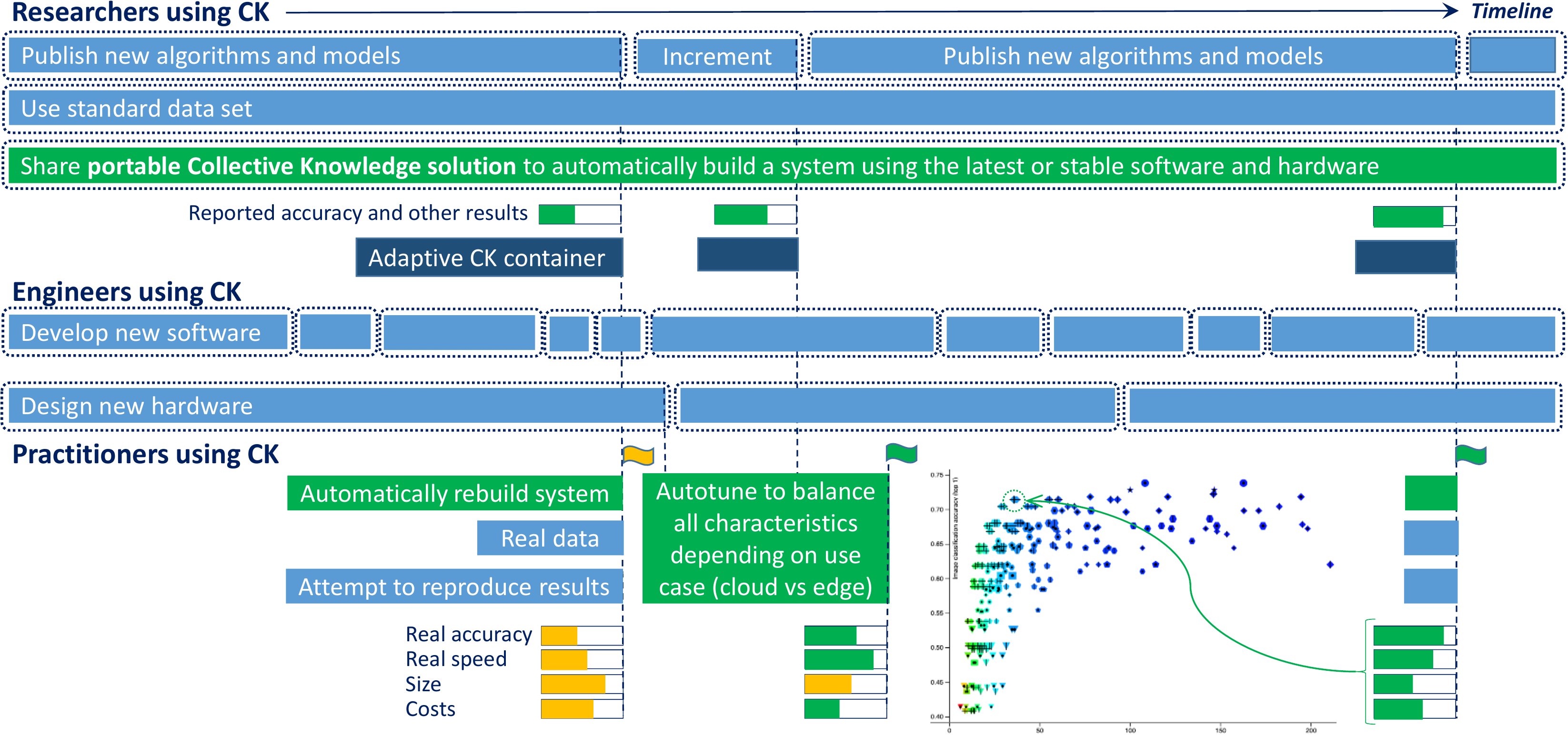}
  \caption{The CK concept helps to connect researchers and practitioners
  to co-design complex computational systems using DevOps principles 
  while automatically adapting to continuously evolving software, hardware, models and datasets.
  The CK framework and public dashboards also help to unify, automate and crowdsource 
  the benchmarking and auto-tuning process across diverse components
  from different vendors to automatically find the most efficient systems 
  on the Pareto frontier.
  }
  \label{fig:timeline2}
\end{figure*}

CK use cases demonstrated that it was possible to develop and use a common research infrastructure 
with different levels of abstraction to bridge the gap between researchers and practitioners 
and help them to collaboratively co-design efficient computational systems.
Scientists could then work with a higher-level abstraction while allowing engineers
to continue improving the lower-level abstractions for continuously evolving software and hardware 
in deploying new techniques in production without waiting for each other, as shown in Figure~\ref{fig:timeline2}.
Furthermore, the unified interfaces and meta descriptions of all CK components and workflows made it possible
to explain what was happening inside complex and "black box" computational systems, 
integrate them with legacy systems, use them inside "adaptive" Docker, and share them along with published papers
while applying the DevOps methodology and agile principles in scientific research.


\section{CK platform}

The practical use of CK as a portable and customizable workflow framework in multiple academic 
and industrial projects exposed several limitations:
  
\begin{itemize} 

 \item The distributed nature of the CK technology, the lack of
 a centralized place to keep all CK components, automation actions and workflows,
 and the lack of a convenient GUI made it very challenging to keep track of all 
 contributions from the community.
 As a result, it is not easy to discuss and test APIs,
 add new components and assemble workflows, automatically validate them 
 across diverse platforms and environments, and connect them with legacy systems.

 \item The concept of backward compatibility of CK APIs and the lack
 of versioning similar to Java made it challenging to keep stable and
 bug-free workflows in the real world: any bug in a reusable CK component 
 from one GitHub project could easily break dependent workflows 
 in another GitHub project.

 \item The CK command-line interface with the access to all automation actions with numerous parameters 
 was too low-level for researchers.
 This is similar to the situation with Git, a powerful but quite complex and CLI-based tool 
 that requires extra web services such as GitHub and GitLab to make it more user-friendly.

\end{itemize}

\begin{figure*}[ht]
  \centering
  \includegraphics[width=1.0\textwidth]{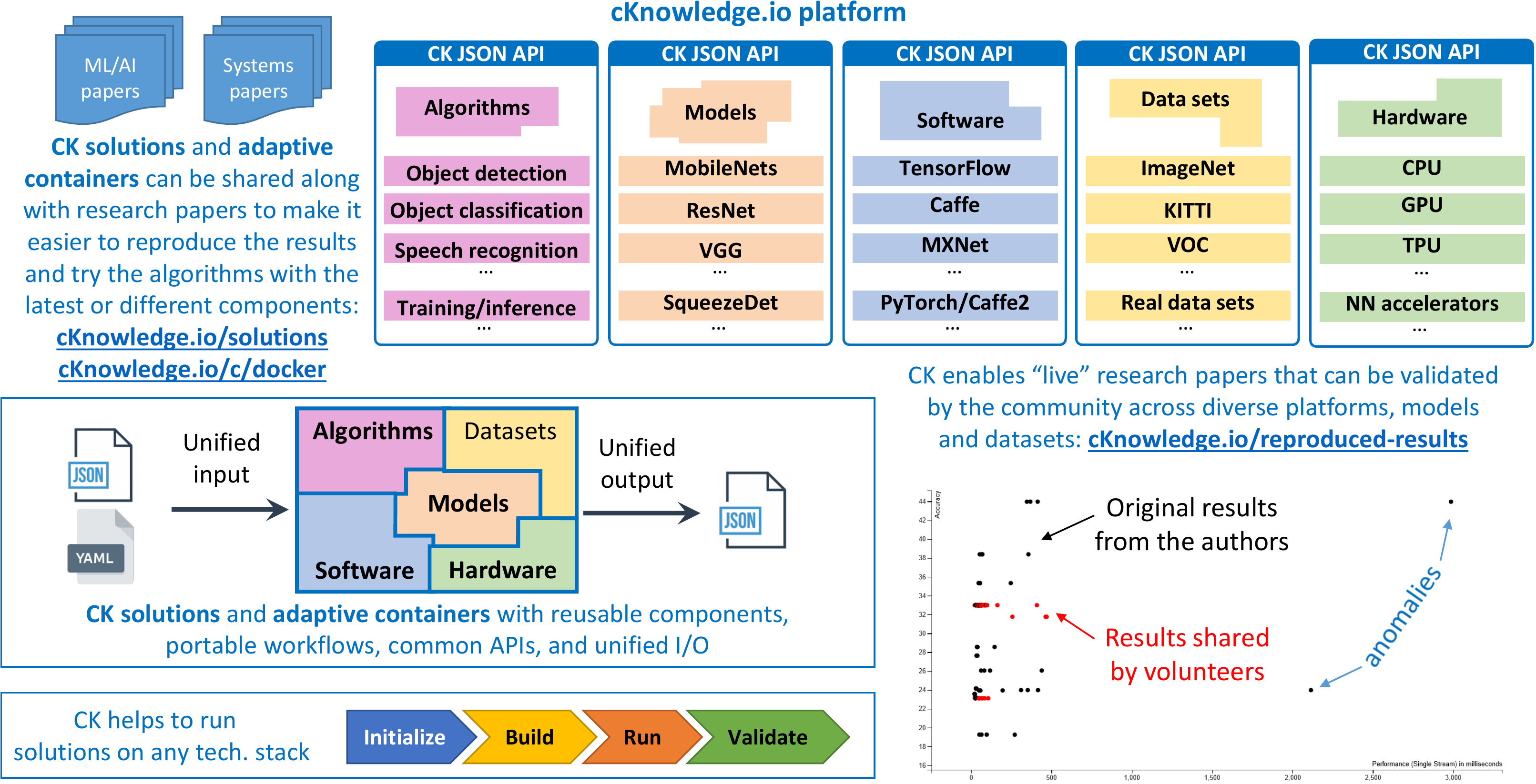}
  \caption{cKnowledge.io: an open platform to organize scientific knowledge in the form of
  portable workflows, reusable components and reproducible research results.
  It already contains many automation actions and components needed to co-design efficient and self-optimizing computational systems,
  enable reproducible and live papers validated by the community, and keep track of the state-of-the-art research techniques
  that can be deployed in production.}
  \label{fig:ck2}
\end{figure*}

This feedback from CK users motivated me to start developing \href{https://cKnowledge.io}{cKnowledge.io} 
(Figure~\ref{fig:ck2}), an open web-based platform with a GUI to aggregate, version
and test all CK components and portable workflows.
I also wanted to replace aging optimization repository at cTuning.org with an extensible and modular platform 
to crowdsource and reproduce tedious experiments such as benchmarking and co-design of efficient systems 
for AI and ML across diverse platforms and data provided by volunteers.

The CK platform is inspired by GitHub and PyPI: I see it as
a collaborative platform to share reusable automation actions for repetitive research tasks
and assemble portable workflows.
It also includes the open-source CK client~\cite{ck-client} that provides a common API to initialise, 
build, run and validate different research projects based on a simple JSON or YAML manifest.
This client is connected with live scoreboards on the CK platform
to collaboratively reproduce and compare the state-of-the-art research results during artifact evaluation 
that we helped to organize at ML and Systems conferences~\cite{reproduced-results}.

My intention is to use the CK platform to complement and enhance 
MLPerf, the ACM Digital Library, PapersWithCode.com, 
and existing reproducibility initiatives and artifact evaluation at ACM, IEEE and NeurIPS conferences
with the help of CK-powered adaptive containers, portable workflows, reusable components, "live" papers 
and reproducible results validated by the community
using realistic data across diverse models, software and hardware.

\section{CK demo: automating and customizing AI/ML benchmarking}

I prepared a live and interactive demonstration of the CK solution that automates 
the MLPerf inference benchmark~\cite{reddi2019mlperf}, connects it with the live CK dashboard
(public optimization repository)~\cite{reproduced-results},
and helps volunteers to crowdsource benchmarking and design space exploration 
across diverse platforms similar to the Collective Tuning Initiative~\cite{Fur2009}
and the SETI@home project: \href{https://cKnowledge.io/test}{cKnowledge.io/test}.

This demonstration shows how to use a unified CK API to automatically build, run 
and validate object detection based on SSD-Mobilenet, TensorFlow and COCO dataset 
across Raspberry Pi computers, Android phones, laptops, desktops and data centers.
This solution is based on a simple JSON file describing the following tasks
and their dependencies on CK components:
\begin{itemize} 
 \item prepare a Python virtual environment (can be skipped for the native installation)
 \item download and install the Coco dataset (50 or 5000 images)
 \item detect C++ compilers or Python interpreters needed for object detection
 \item install Tensorflow framework with a specified version for a given target machine
 \item download and install the SSD-MobileNet model compatible with selected Tensorflow
 \item manage installation of all other dependencies and libraries
 \item compile object detection for a given machine and prepare pre/post-processing scripts
\end{itemize}

This solution was published on the \href{https://cKnowledge.io}{cKnowledge.io} platform 
using the open-source CK client~\cite{ck-client} to help users participate in crowd-benchmarking 
on their own machines as follows:

\begin{strip}
\begin{lstlisting}[language=bash, basicstyle=\footnotesize]

# Install the CK client from PyPi using:
pip install cbench
    
# Download and build the solution on a given machine (example for Linux): 
cb init demo-obj-detection-coco-tf-cpu-benchmark-linux-portable-workflows
    
# Run the solution on a given machine:
cb benchmark demo-obj-detection-coco-tf-cpu-benchmark-linux-portable-workflows

\end{lstlisting}
\end{strip}

The users can then see the benchmarking results (speed, latency, accuracy and other exposed characteristics through the CK workflow)
on the live CK dashboard associated with this solution and compare them against the official MLPerf 
results or with the results shared by other users: 
\href{https://cKnowledge.io/demo-result}{cKnowledge.io/demo-result}.

After validating this solution on a given platform, the users can also clone it and update the JSON description
to retarget this benchmark to other devices and operating systems such as MacOS, Windows, 
Android phones, servers with CUDA-enabled GPUs and so on.

The users have the possibility to integrate such ML solutions with production systems with the help of unified CK APIs
as demonstrated by connecting the above CK solution for object detection with the webcam in any browser:
\href{https://cKnowledge.io/demo-solution}{cKnowledge.io/demo-solution}.

Finally, it is possible to use containers with CK repositories, workflows and common APIs as follows:

\begin{strip}
\begin{lstlisting}[language=bash, basicstyle=\footnotesize]

docker run ctuning/cbrain-obj-detection-coco-tf-cpu-benchmark-linux-portable-workflows \
 /bin/bash -c "cb benchmark demo-obj-detection-coco-tf-cpu-benchmark-linux-portable-workflows

docker run ctuning/cbench-mlperf-inference-v0.5-detection-openvino-ssd-mobilenet-coco-500-linux \
  /bin/bash -c "cb benchmark mlperf-inference-v0.5-detection-openvino-ssd-mobilenet-coco-500-linux

\end{lstlisting}
\end{strip}

Combining Docker and portable CK workflows enables "adaptive" CK containers for MLPerf that can be easily customized, 
rebuilt with different ML models, datasets, compilers, frameworks and tools encapsulated inside CK components,
and deployed in production~\cite{ck-adaptive-containers}.

\section{Conclusions and future work}

My very first research project to prototype an analog neural network stalled in the late 1990s
because it took me far too long to build from scratch all the infrastructure 
to model and train Hopfield neural networks, generate diverse datasets, 
co-design and optimize software and hardware, run and reproduce all experiments,
compare them with other techniques from published papers, 
and use this technology in practice in a completely different environment.

In this article, I explain why I have developed the Collective Knowledge framework
and how it can help to address the issues above by organizing all research projects 
as a database of reusable components, portable workflows
and reproducible experiments based on FAIR principles 
(findable, accessible, interoperable, and reusable).
I also describe how the CK framework attempts to bring DevOps and "Software 2.0" principles to scientific research
and to help users share and reuse best practices, automation actions and research artifacts
in a unified way alongside reproducible papers.
Finally, I demonstrate how the CK concept helps to complement, unify and interconnect existing tools, platforms, and reproducibility initiatives
with common APIs and extensible meta descriptions rather than rewriting them or competing with them.

I present several use cases of how CK helps to connect researchers and practitioners 
to collaboratively design more reliable, reproducible and efficient computational systems 
for machine learning, artificial intelligence and other emerging workloads 
that can automatically adapt to continuously evolving software, hardware, models and datasets.
I also describe the \href{cKnowledge.io}{https://cKnowledge.io} platform 
that I have developed to organize knowledge particularly about AI, ML, systems and other innovative technology 
in the form of portable CK workflows, automation actions, reusable artifacts 
and reproducible results from research papers.
My goal is to help the community to find useful methods from research papers,
quickly build them on any tech stack, integrate them with new or legacy systems,
start using them in the real world with real data, and combine
Collective Knowledge and AI to build better AI systems.

Finally, I demonstrate the concept of "live" research papers connected with portable CK workflows and online CK dashboards
to let the community automatically validate and update experimental results even after the project,
detect and share unexpected behaviour, and fix problems collaboratively~\cite{cm:29db2248aba45e59:c4b24bff57f4ad07}.
I believe that such a collaborative approach can make computational research 
more reproducible, portable, sustainable, explainable and trustable.

However, CK is still a proof-of-concept and there remains a lot to simplify and improve.
Future work will intend to make CK more user-friendly and to simplify the onboarding process, as well as
to standardise all APIs and JSON meta descriptions. It will also focus on development of a simple GUI to create and share automation actions and CK components, 
assemble portable workflows, run experiments, compare research techniques, generate adaptive containers,
and participate in lifelong AI, ML and systems optimization.

My long-term goal is to use CK to develop a virtual playground, an optimization repository and a marketplace 
where researchers and practitioners assemble AI, ML and other novel applications similar to live species 
that continue to evolve, self-optimize and compete with each other across diverse tech stacks from different vendors and users.
The winning solutions with the best trade-offs between speed, latency, accuracy, energy, size, costs 
and other metrics can then be selected at any time from the Pareto frontier based on user requirements and constraints.
Such solutions can be immediately deployed in production on any platform from data centers to the edge 
in the most efficient way thus accelerating AI, ML and systems innovation and digital transformation.


\section*{Acknowledgements}

I thank 
Sam Ainsworth, Erik Altman, Lorena Barba, Victor Bittorf, Unmesh D. Bordoloi, Steve Brierley, Luis Ceze, Milind Chabbi,
Bruce Childers, Nikolay Chunosov, Marco Cianfriglia, Albert Cohen, Cody Coleman, Chris Cummins, Jack Davidson,
Alastair Donaldson, Achi Dosanjh, Thibaut Dumontet, Debojyoti Dutta, Daniil Efremov, Nicolas Essayan, Todd Gamblin,
Leo Gordon, Wayne Graves, Christophe Guillon, Herve Guillou, Stephen Herbein, Michael Heroux, Patrick Hesse,
James Hetherignton, Kenneth Hoste, Robert Hundt, Ivo Jimenez, Tom St. John, Timothy M. Jones, David Kanter,
Yuriy Kashnikov, Gaurav Kaul, Sergey Kolesnikov, Shriram Krishnamurthi, Dan Laney, Andrei Lascu, Hugh Leather,
Wei Li, Anton Lokhmotov, Peter Mattson, Thierry Moreau, Dewey Murdick, Luigi Nardi, Cedric Nugteren, Michael O'Boyle,
Ivan Ospiov, Bhavesh Patel, Gennady Pekhimenko, Massimiliano Picone, Ed Plowman, Ramesh Radhakrishnan, Ilya Rahkovsky,
Vijay Janapa Reddi, Vincent Rehm, Alka Roy, Shubhadeep Roychowdhury, Dmitry Savenko, Aaron Smith, Jim Spohrer,
Michel Steuwer, Victoria Stodden, Robert Stojnic, Michela Taufer, Stuart Taylor, Olivier Temam, Eben Upton,
Nicolas Vasilache, Flavio Vella, Davide Del Vento, Boris Veytsman, Alex Wade, Pete Warden, Dave Wilkinson,
Matei Zaharia, Alex Zhigarev 
and other great colleagues for interesting discussions, 
practical use cases and useful feedback.


\bibliographystyle{plain}
\bibliography{ck}

\end{document}